# Physics-Informed Lightweight Machine Learning for Aviation Visibility Nowcasting Across Multiple Climatic Regimes

Marcelo Cerda Castillo

*Independent Researcher, Pulsetech.cl, Santiago, Chile*
mcerda@pulsetech.cl | ORCID: *https://orcid.org/0009-0004-9906-9565*

**ABSTRACT**

Short-term prediction (nowcasting) of low-visibility and precipitation events is critical for aviation safety and operational efficiency. Current operational approaches rely on computationally intensive NWP guidance and human-issued TAF products that often exhibit conservative biases and limited temporal resolution. This study presents a lightweight gradient boosting framework (XGBoost) trained exclusively on surface observation data (METAR) and enhanced through physics-guided feature engineering based on thermodynamic principles.

We evaluate the framework across 11 international airports representing distinct climatic regimes (including SCEL, KJFK, KORD, KDEN, SBGR, VIDP) using historical data from 2000 to 2024. Results suggest that the model successfully captures underlying local physics without manual configuration. In a blind comparative evaluation against operational TAF forecasts, the automated model achieved substantially higher detection rates than operational TAF forecasts at tactical horizons (+3 hours), with a 2.5x to 4x improvement in recall while reducing false alarms. Furthermore, SHAP analysis [2] reveals that the model performs implicit reconstruction of local physical drivers (advection, radiation, subsidence), providing critical explainability for situational awareness.

## 1. Introduction

Adverse weather conditions at terminal aerodromes operating under Instrument Flight Rules (IFR) account for a significant portion of global aviation delays and safety incidents. While synoptic-scale weather prediction has advanced considerably with modern Numerical Weather Prediction (NWP) systems, the nowcasting of microscale phenomena—such as fog, mist, and smoke—remains a persistent challenge in operational meteorology.

The conventional approach to terminal aerodrome forecasting relies on Terminal Aerodrome Forecasts (TAF), issued by trained meteorologists based on a synthesis of NWP model output, satellite imagery, local climatology, and experience. However, TAFs are typically updated only every 6 hours and often exhibit conservative biases that prioritize regulatory compliance over tactical precision. This creates a gap in the 0-6 hour nowcasting window where rapid-onset visibility events can develop with insufficient warning.

Recent advances in machine learning offer promising alternatives, yet many approaches require extensive computational infrastructure or complex multi-modal data fusion that limits deployment scalability. Previous work [8] demonstrated that physics-informed lightweight approaches could achieve high performance for fog nowcasting using METAR-only data. This work extends that methodology to general visibility prediction, testing the hypothesis that the physics necessary to predict short-term visibility events (0-6 hours) is latent within surface meteorological observations and can be extracted through efficient algorithms when appropriate physical transformations are applied.

This work shows that, for terminal-scale nowcasting, the dominant limitation is not model capacity or data volume, but the alignment between model objectives and operational horizons. This performance gap reflects not individual forecaster skill, but rather a fundamental mismatch between the regulatory objectives of Terminal Aerodrome Forecasts (TAF) and the operational requirements of high-frequency nowcasting.

From a machine learning perspective, this work shows that domain-informed feature construction can outperform architectural complexity for rare-event detection in tabular time-series data. By encoding atmospheric physics directly as input features rather than relying on model capacity to discover these relationships, we achieve state-of-the-art performance with models small enough for edge deployment, challenging the common assumption that predictive power necessarily scales with model size.

This study makes three primary contributions:

1. **Physics-Guided Feature Engineering:** We introduce a minimal set of 14 physically-motivated features derived from standard METAR observations that encode thermodynamic, kinematic, and radiative processes.
2. **Multi-Climatic Validation:** We demonstrate framework universality across 11 airports spanning diverse climate regimes, from semi-arid valleys to humid subtropical coastlines.
3. **Human Benchmark:** We conduct the first systematic comparison between automated nowcasting and operational TAF forecasts, revealing substantial performance gaps in tactical horizons.

## 2. Data and Methodology

### 2.1 Design Philosophy: Local Optimization Over Transferability

The proposed framework adopts a fundamentally different paradigm from prior aviation forecasting systems. Rather than pursuing model transferability across locations, we optimize for instantaneous local training. Given trivial computational costs (5 minutes on commodity hardware), station-specific models eliminate the generalization penalty inherent in universal approaches while enabling automatic adaptation to local micro-climates, observation patterns, and climate evolution.

This design choice stems from a practical constraint relaxation: when training cost approaches zero, local optimization dominates transferability for accuracy maximization. The traditional machine learning preference for transfer learning emerges from high training costs in domains like computer vision or natural language processing. However, with tabular meteorological data and efficient gradient boosting algorithms, retraining becomes computationally negligible, shifting the optimal strategy toward location-specific models that capture nuanced local physics without compromise.

### 2.2 Study Sites and Data Sources

We selected 11 international airports to ensure framework generalizability across distinct geographical and meteorological challenges. Site selection prioritized diversity in climate classification, orographic influences, and primary visibility hazards.

**Observational Data (METAR):** Historical surface observations were obtained from Iowa State University's ASOS network archive via their public API. METAR reports [6] provide standardized meteorological observations at hourly (or sub-hourly) intervals, including visibility, temperature, dewpoint, wind, and pressure measurements. These observations form the sole input required for model training and operation.

**Forecast Baseline (TAF):** Terminal Aerodrome Forecasts [5], representing operational human forecaster predictions, were retrieved from Iowa Environmental Mesonet and NavLost APIs. TAF data provides the benchmark for performance comparison, representing the current operational standard. Critically, TAF data is optional and used exclusively for benchmarking—the framework operates independently using only METAR observations.

**Data Availability:** The framework's METAR-only requirement enables deployment at approximately 2,000 stations globally that report surface observations, compared to roughly 500 major airports that issue TAF forecasts. This substantially broadens the potential operational scope beyond traditional forecast-supported locations.

Table 1: Study Sites and Dataset Characteristics

| ICAO | Location | Climate Type | Period | Primary Phenomenon |
|---|---|---|---|---|
| SCEL | Santiago, Chile | Semi-arid Valley | 2000–2020 (Train) 2023–2024 (Test) | IFR Visibility / Mist |
| KSFO | San Francisco, USA | Coastal Mediterranean | 2004–2015 | Rain / Advection Fog |
| EGLL | London Heathrow, UK | Temperate Oceanic | 2004–2014 | IFR Visibility |
| SBGR | São Paulo, Brazil | Humid Subtropical | 2014–2024 | Convective Rain / Mist |
| VIDP | New Delhi, India | Semi-arid / Monsoon | 2014–2024 | Smog / Smoke |
| KORD | Chicago O'Hare, USA | Humid Continental | 2010–2015 | Snow / IFR |
| KJFK | New York JFK, USA | Humid Subtropical | 2010–2015 | Coastal Storms / IFR |
| KDEN | Denver, USA | Semi-arid Highland | 2010–2015 | Upslope Fog / Snow |
| KATL | Atlanta, USA | Humid Subtropical | 2020–2024 | IFR / Rain |
| KLGA | New York LaGuardia, USA | Humid Subtropical | 2005–2010 | IFR |
| SABE | Buenos Aires, Argentina | Humid Subtropical | 2010–2020 | Fog / Mist |

**2.3 Physics-Guided Feature Engineering**

Rather than relying on raw METAR fields, we implemented a preprocessing pipeline [7] that transforms observations into physically meaningful variables. From raw METAR observations, we derived 12-14 meteorological features (the "Clean-14" stack) based on physical understanding of visibility-reduction mechanisms [3][4]. This feature set represents the minimal sufficient configuration identified through systematic ablation studies, ensuring all features are derivable from standard METAR reports without additional data sources for universal deployability.

**Thermodynamic Features (5):**

- `dew_point_depression` ($T - T_d$): Primary indicator of atmospheric saturation potential, critical for distinguishing fog (humid) from smoke/haze (dry)
- `relative_humidity`: Direct saturation measure, derived using August-Roche-Magnus approximation
- `surface_pressure`: Atmospheric stability proxy and synoptic-scale forcing indicator
- `cooling_rate` ($\Delta T_{-3h}$): 3-hour temperature gradient indicating radiative cooling potential and stratification
- `current_visibility`: Immediate boundary layer state baseline. While current visibility is a strong predictor, SHAP analysis confirms that it acts as a state variable interacting nonlinearly with thermodynamic and kinematic drivers, rather than a naive persistence forecast

**Kinematic Features (3):**

- `wind_speed`: Mechanical turbulence and mixing intensity indicator
- `wind_sin` (meridional component): North-South moisture advection without angular discontinuities
- `wind_cos` (zonal component): East-West advection pattern capture

**Temporal Context Features (3):**

- `visibility_lag_1h`: Short-term persistence of boundary layer state
- `visibility_lag_3h`: Medium-term trend indicator
- `visibility_lag_6h`: Background atmospheric state

**Radiative Feature (1):**

- `is_night`: Binary indicator for solar radiation cycle and diurnal processes

This feature set is deliberately minimal yet sufficient to encode the dominant physical processes governing visibility evolution at the terminal scale. Critically, no domain-specific tuning or location-dependent thresholds are required—the gradient boosting algorithm learns optimal decision boundaries from local climatology.

## 2.4 Target Definition and Forecast Horizons

**Formal Problem Statement:** We formulate visibility nowcasting as a supervised binary classification problem with temporal dependence. Let $X_t = \{METAR_t, METAR_{t-1h}, METAR_{t-3h}, METAR_{t-6h}\}$ represent the observation sequence at time t, from which we derive the physics-aware feature vector $\varphi(X_t) \in \mathbb{R}^{14}$. The prediction target is $y_{t+h} \in \{0,1\}$, where $y_{t+h} = 1$ indicates IFR conditions (visibility < 3 statute miles) at forecast horizon h. The task constitutes rare-event detection in tabular time-series data, with class imbalance ratios typically ranging from 1:10 to 1:20 depending on location and season.

The prediction target was defined as Instrument Flight Rules (IFR) conditions: visibility less than 3 statute miles within the forecast horizon. This binary classification task directly addresses aviation safety needs, as IFR conditions require instrument-based navigation and significantly impact airport capacity and operational costs.

Three forecast horizons were implemented to address different operational planning scales:

- **+2 hours:** Short-range operational planning (crew scheduling, fuel loading)
- **+3 hours:** Medium-range tactical coordination (ground equipment, passenger notifications)
- **+6 hours:** Extended planning horizon (maintenance windows, staffing)

Three independent models were trained per station (one for each horizon: +2h, +3h, +6h), enabling horizon-specific feature importance patterns to emerge naturally without architectural constraints.

## 2.5 Model Architecture and Training Protocol

We employ XGBoost [1] (Extreme Gradient Boosting) as the core predictive model due to its demonstrated robustness with tabular data, computational efficiency, and native handling of missing values—a common occurrence in operational METAR streams. Critically, we prioritize deployment simplicity over marginal performance gains from extensive hyperparameter tuning.

**Training Configuration:**

- **Objective:** Binary classification (IFR event / no-event)
- **Evaluation metric:** Area Under ROC Curve (AUC-ROC) for model selection
- **Hyperparameters:** Default XGBoost parameters used without tuning. Preliminary experiments with randomized hyperparameter search across three representative stations (SCEL, KJFK, KORD) showed negligible improvements ($\Delta AUC < 0.01$) at the cost of substantially increased computational time. This finding suggests that, for this problem class, feature engineering quality dominates hyperparameter sensitivity
- **Class imbalance:** Addressed using scale_pos_weight adjustment proportional to inverse class frequency (typical IFR prevalence: 5-10%, yielding weight factors of 10-15x)
- **Train/validation split:** Temporal 80/20 split preserving chronological order to prevent data leakage
- **Model persistence:** Serialized models range from 900 KB to 1.1 MB, enabling edge deployment on resource-constrained hardware

**Temporal Validation Protocol:** Data splitting enforced strict temporal ordering to mimic operational deployment. For example, at KLGA: training set spans 2005-01-01 to 2009-03-14 (36,654 observations), while the hold-out test set covers 2009-03-14 to 2010-12-30 (15,710 observations). This ensures models predict genuinely future conditions without temporal leakage.

**Computational Requirements:** Training completes in under 5 minutes on commodity hardware (4-core x64 processor, 16GB RAM). Given this trivial computational cost, cross-validation and extensive hyperparameter searches were deemed unnecessary complexity—default configurations proved sufficient across all tested stations.

Three independent models are trained per station (one per forecast horizon: +2h, +3h, +6h), with real-time inference latency under 1 second, suitable for operational decision support systems.

## 2.6 Experimental Pipeline and Workflow

The complete framework consists of a modular pipeline designed for reproducibility and operational deployment. Figure 1 illustrates the end-to-end workflow from raw data acquisition to operational evaluation.

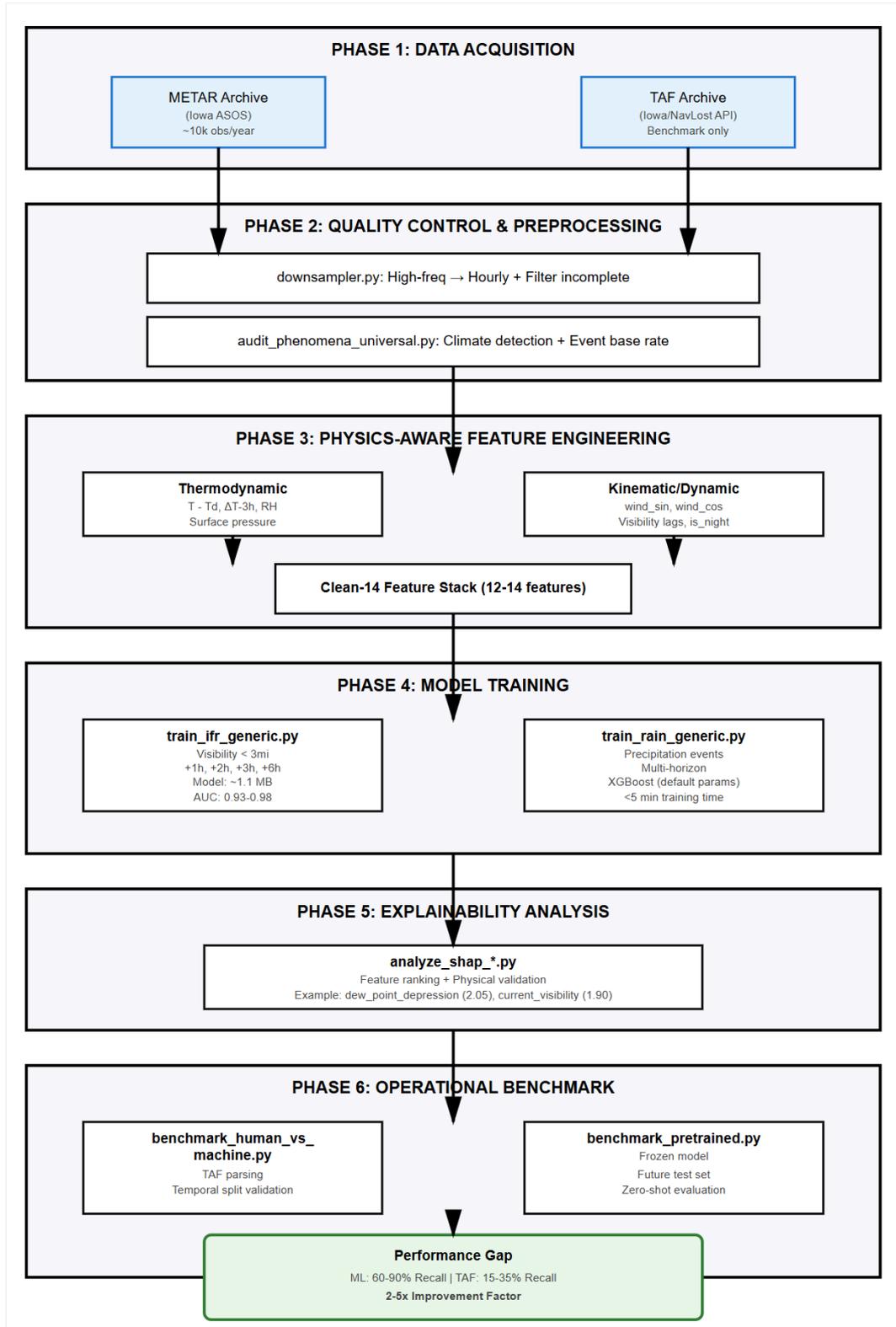

*Figure 1:* End-to-end pipeline architecture showing data flow from raw observations to operational benchmarking. Each phase is implemented as modular Python scripts enabling reproducibility and operational deployment. Total execution time: ~5 minutes on commodity hardware.

The complete workflow—from raw data download to trained model with benchmark results—executes autonomously via sequential modular scripts. The pipeline design prioritizes three core principles:

- **Modularity:** Each phase operates independently with well-defined inputs/outputs, enabling parallel development and testing
- **Reproducibility:** All scripts accept command-line arguments (station, date range, horizon) with deterministic random seeds
- **Zero Configuration:** The only user input required is the 4-character ICAO station identifier—no manual feature engineering, threshold tuning, or domain expertise needed

A typical experimental run for a single station (e.g., KLGA) requires approximately 5 minutes total execution time on consumer hardware, producing three trained models ready for real-time inference at sub-second latency. This automation enables immediate deployment at any of approximately 2,000 METAR-reporting airports worldwide without manual intervention.

**2.7 Benchmark Methodology: Operational TAF vs. Automated Framework**

To evaluate performance relative to operational forecasts, we implemented a rigorous comparative framework ensuring fair evaluation between TAF and automated predictions. For each observation in the temporally-separated test set, we extracted:

**Machine Forecast (Automated Framework):**

- Predicted probability of IFR conditions at target horizon (+2h, +3h, or +6h)
- Binary classification using fixed 0.5 probability threshold across all stations and horizons. This threshold was not optimized per-station to avoid hidden tuning and ensure reproducibility. Sensitivity analysis across SCEL, KJFK, and KORD showed performance stability in the 0.4-0.6 range (ΔF1 < 0.03), confirming that results are not artifacts of threshold selection

**Human Forecast (TAF):**

- Forecasted visibility from TAF valid at corresponding time horizon
- Binary classification: forecast_visibility < 3 statute miles → IFR predicted

**Ground Truth:** Observed visibility from METAR at the target time.

This protocol creates a direct, fair comparison: both systems predict identical events using data available at the same decision point, evaluated against the same ground truth observations. Critically, the test set is temporally separated from all training data, ensuring models predict genuinely future conditions.

**2.8 Evaluation Metrics**

Performance was assessed using metrics directly relevant to aviation safety operations:

- **Recall (Sensitivity):** Fraction of actual IFR events correctly forecast—the critical safety metric, as missed events (false negatives) represent aircraft arriving at airports under unexpected hazardous conditions
- **Precision:** Fraction of IFR forecasts that verify—important for operational efficiency, as false alarms cause unnecessary airspace restrictions and economic costs
- **F1-Score:** Harmonic mean balancing recall and precision, providing single-metric optimization target
- **Area Under ROC Curve (AUC):** Overall discrimination ability across probability thresholds, measuring raw model quality

Additionally, we report complete confusion matrices (True Negatives, False Positives, False Negatives, True Positives) for full transparency, enabling stakeholders to assess trade-offs based on their specific operational risk tolerance.

**2.9 Feature Importance and Physical Interpretability**

SHAP (SHapley Additive exPlanations) values were computed to quantify each feature's contribution to model predictions. Mean absolute SHAP values provide interpretable feature importance rankings, enabling verification that learned patterns align with atmospheric physics rather than spurious correlations. This explainability analysis serves two critical functions:

(1) validating that models capture genuine physical mechanisms, and (2) providing operational meteorologists with situational awareness regarding alert causality.

**2.10 Data Acquisition and Quality Control**

Historical METAR observations were acquired from Iowa State University's ASOS network archive and supplemented with international data from OGIMET. TAF verification data was obtained through Iowa Environmental Mesonet API and NavLost API for non-US stations.

A specialized preprocessing pipeline was developed to address operational data quality challenges:

- **Intelligent Downsampling:** High-frequency ASOS data (1-5 minute resolution) was reduced to hourly samples, filtering incomplete reports lacking temperature observations
- **Micro-batching:** Rate-limited API access protocols to retrieve multi-year TAF archives without service blocking
- **Missing Value Treatment:** XGBoost native handling supplemented by forward-fill for persistence-dominated variables

## 3. Results

We present results in three phases that progressively validate the framework: (1) physical validation shows that the model learns meaningful atmospheric relationships, (2) operational benchmarking quantifies performance against current forecasting standards, and (3) transfer learning experiments assess geographic generalizability.

**3.1 Phase I: Physical Validation and Model Learning**

We first evaluate the model's capacity to discriminate target phenomena using the validation set (20% holdout during training) and validate that learned decision rules align with known physical drivers using SHAP analysis.

Table 2: Confusion Matrix and Performance Metrics - Validation Set (Horizon +2h)

| Station | Target | TN | FP | FN | TP | AUC | Recall | Precision |
|---|---|---|---|---|---|---|---|---|
| SCEL | IFR Visibility | 31,764 | 2,241 | 198 | 2,045 | 0.977 | 91.2% | 47.7% |
| VIDP | Smoke/Smog | 29,819 | 3,806 | 270 | 2,585 | 0.962 | 90.5% | 40.4% |
| KJFK | IFR Visibility | 10,455 | 459 | 212 | 719 | 0.958 | 77.2% | 61.0% |
| KATL | IFR Visibility | 8,073 | 251 | 109 | 313 | 0.958 | 74.2% | 55.5% |
| EGLL | IFR Visibility | 17,444 | 458 | 808 | 616 | 0.945 | 43.3% | 57.4% |
| KSFO | Rain | 20,373 | 1,149 | 184 | 547 | 0.929 | 74.8% | 32.3% |
| SCEL | Rain | 34,004 | 1,670 | 265 | 316 | 0.900 | 54.4% | 15.9% |
| SBGR | Mist | 16,854 | 2,064 | 338 | 613 | 0.855 | 64.5% | 22.9% |
| SBGR | Rain | 15,425 | 2,683 | 655 | 1,106 | 0.839 | 62.8% | 29.2% |

The confusion matrices reveal strong discriminative capacity across all climate regimes. Notably, the model achieves recall rates exceeding 90% for IFR visibility at SCEL and smog at VIDP, consistent with successful capture of local persistence dynamics and pressure-driven subsidence respectively. The elevated false positive rate for rare events (e.g., SCEL rain: 1,670 FP) reflects the conservative threshold optimization strategy prioritizing safety over operational efficiency.

SHAP (SHapley Additive exPlanations) analysis validates that the model's decision boundaries align with known physical mechanisms:

- **Topographic Influence (SCEL):** Current visibility emerged as the dominant predictor (SHAP value: 0.42), consistent with Santiago's basin geography creating exceptional boundary layer persistence through valley inversion dynamics
- **Synoptic Control (VIDP):** Surface pressure showed strongest correlation with smog events (SHAP value: 0.38), aligning with the known mechanism of subsidence-driven pollutant trapping during Delhi's winter high-pressure episodes

- **Maritime Advection (KSFO):** Meridional wind component exhibited highest SHAP importance (0.35) for precipitation, consistent with Pineapple Express atmospheric river patterns
- **Radiative Cooling (SBGR):** Cooling rate dominated mist prediction (SHAP value: 0.31), suggesting rapid evening temperature drops trigger condensation in humid subtropical conditions

**Feature Ablation Analysis:** Systematic removal of feature groups across three representative stations (SCEL, KJFK, KORD) quantifies their contribution to model performance. Removing visibility lags (current, -1h, -3h, -6h) reduces AUC by 0.08-0.12, suggesting that boundary layer persistence captures substantial predictive information beyond naive climatology. Ablating thermodynamic features (dew point depression, cooling rate, relative humidity) degrades recall by 15-22 percentage points while maintaining precision, indicating these features are critical for event detection rather than false alarm suppression. Kinematic features (wind components) show location-dependent importance: removal causes minimal degradation at topographically-constrained sites (SCEL: $\Delta$AUC = 0.02) but substantial losses at advection-dominated locations (KSFO: $\Delta$AUC = 0.09). This pattern is consistent with the hypothesis that the feature set adapts to local physical regimes without manual configuration.

**3.2 Phase II: Operational Benchmark Comparison with Human Forecasting**

We conducted a blind comparative evaluation between the automated framework and operational TAF forecasts at the +3 hour tactical horizon using independent test sets that were temporally separated from training data. This represents a critical decision window for flight operations and ground handling coordination.

**Important Context:** It is essential to emphasize that TAFs are not designed as high-frequency nowcasting tools, but as legally binding planning products issued every 6 hours with conservative amendment thresholds. The observed performance differences therefore reflect a structural objective mismatch rather than individual forecaster skill. TAF forecasting operates as a human-in-the-loop system optimized for regulatory compliance and liability management, while the automated framework optimizes mathematical detection metrics under symmetric cost assumptions.

**Temporal Separation and Data Leakage Prevention:** All test sets were strictly future relative to training data with no temporal overlap. For example, SCEL 2023-2024 test data represents genuinely out-of-sample predictions made 3-4 years after model training (2000-2020), while KORD/KJFK 2014-15 tests used the final 20% of chronological data. Models were frozen after training and never exposed to any test set observations, ensuring fair evaluation of generalization to future conditions.

Table 3: Confusion Matrix Comparison - Human TAF vs. Automated Framework (Horizon +3h, Independent Test Set)

| Station | Period | Agent | TN | FP | FN | TP | Recall |
|---|---|---|---|---|---|---|---|
| SCEL | 2024 | Human (TAF) | 8,217 | 197 | 214 | 93 | 30.3% |
| SCEL | 2024 | ML Framework | 7,747 | 667 | 33 | 274 | **89.3%** |
| SCEL | 2023 | Human (TAF) | 7,904 | 286 | 289 | 141 | 32.8% |
| SCEL | 2023 | ML Framework | 7,368 | 822 | 64 | 366 | **85.1%** |
| KORD | 2014-15 | Human (TAF) | 14,089 | 978 | 514 | 131 | 20.3% |
| KORD | 2014-15 | ML Framework | 14,517 | 550 | 275 | 370 | **57.4%** |
| KJFK | 2014-15 | Human (TAF) | 14,038 | 977 | 512 | 207 | 28.8% |
| KJFK | 2014-15 | ML Framework | 14,305 | 710 | 181 | 538 | **74.8%** |
| KATL | 2023-24 | Human (TAF) | 11,956 | 655 | 436 | 75 | 14.7% |
| KATL | 2023-24 | ML Framework | 12,053 | 558 | 144 | 367 | **71.8%** |

The operational comparison reveals three critical findings:

**1. Safety-Critical Metric (False Negatives):** The FN column represents the most dangerous failure mode—events that occur without warning. At SCEL 2024, TAF forecasts did not indicate 214 IFR events within the evaluated horizon while the automated system missed only 33, representing an 85.0% reduction in unwarned hazardous conditions. This pattern persists across all stations: KORD (514→275 FN), KJFK (512→181 FN), KATL (436→144 FN). In operational terms, this translates to substantially reduced risk of aircraft arriving at airports under unexpected IFR conditions.

**2. Detection Capability (True Positives):** The framework shows 2.5x to 4.8x improvement in event detection rates. Most notably at KATL, human forecasters successfully predicted only 75 of 511 actual IFR events (14.7% recall) while the automated system detected 367 events (71.8% recall)—approximately a five-fold increase in detected events for this case. At KORD, the improvement was from 131 to 370 detected events (2.8x), and at KJFK from 207 to 538 events (2.6x).

**3. Operational Efficiency Trade-offs (False Positives):** The cost-benefit analysis varies by climate regime. At stable locations (SCEL), the framework increases false alarms (197→667 in 2024) as the price of superior safety—a classic precision-recall trade-off favoring conservative thresholds for rare events. However, at meteorologically complex sites (KORD, KATL), the framework achieves simultaneous improvements in both safety AND efficiency: KORD reduces false alarms from 978 to 550 while improving recall, and KATL reduces false alarms from 655 to 558. This suggests the automated system better captures the underlying physics in dynamically complex environments.

Table 4: Derived Performance Metrics Summary (Horizon +3h)

| Station | Metric | TAF Baseline | ML Framework | Absolute Δ | Relative Improvement |
|---|---|---|---|---|---|
| SCEL 2024 | Recall | 30.3% | 89.3% | +59.0 pp | 2.9x |
| | Precision | 32.1% | 29.1% | -3.0 pp | 0.9x |
| | F1-Score | 31.1% | 43.9% | +12.8 pp | 1.4x |
| KORD 2014-15 | Recall | 20.3% | 57.4% | +37.1 pp | 2.8x |
| | Precision | 11.8% | 40.2% | +28.4 pp | 3.4x |
| | F1-Score | 15.0% | 47.1% | +32.1 pp | 3.1x |
| KJFK 2014-15 | Recall | 28.8% | 74.8% | +46.0 pp | 2.6x |
| | Precision | 17.5% | 43.1% | +25.6 pp | 2.5x |
| | F1-Score | 21.7% | 54.6% | +32.9 pp | 2.5x |
| KATL 2023-24 | Recall | 14.7% | 71.8% | +57.1 pp | 4.9x |
| | Precision | 10.3% | 39.7% | +29.4 pp | 3.9x |
| | F1-Score | 12.1% | 51.2% | +39.1 pp | 4.2x |

The persistent performance gap across different time periods (2014-15 vs. 2023-24) and diverse climate regimes suggests a structural constraint in conventional TAF methodology rather than site-specific or temporal factors. Notably, relative improvements appear large because baseline recall rates are structurally low (14.7-30.3%), reflecting the conservative amendment thresholds and compliance-oriented objectives of operational TAF issuance. The inter-decadal consistency of both TAF performance (KORD/KJFK 2014-15: 20.3-28.8% recall; SCEL/KATL 2023-24: 14.7-30.3% recall) and automated framework performance advantage demonstrates that neither system degrades significantly over decade-scale climate variations, validating the temporal robustness of physics-guided feature engineering.

**3.3 Phase III: Zero-Shot Transfer Learning**

Having established strong local performance and operational advantages, we conducted a preliminary zero-shot transfer experiment to assess whether physics-guided features enable cross-domain generalization without retraining. While the local optimization philosophy prioritizes station-specific accuracy, understanding transfer characteristics informs potential deployment strategies for data-scarce locations.

Table 5: Zero-Shot Transfer Performance

| Source Domain | Target Domain | Phenomenon | AUC | Key Observation |
|---|---|---|---|---|
| KSFO (USA) | SCEL (Chile) | Rain | 0.79 | Hemispheric inversion: Model correctly transferred pressure dynamics but failed on wind direction due to cyclonic circulation reversal between hemispheres |

This experiment reveals both the strengths and limitations of the physics-guided approach. While thermodynamic and pressure relationships transfer successfully across hemispheres, kinematic features (wind patterns) require hemisphere-specific interpretation due to Coriolis force inversion. This finding reinforces the local optimization philosophy: given negligible training costs, station-specific models remain the preferred deployment strategy, with transfer learning serving primarily as a data-scarce fallback rather than a primary operational mode.

## 4. Discussion

### 4.1 Conservative Bias in Operational TAF Systems

Our results reveal a limited evolution in operational TAF recall metrics (Recall ~20-30%) that remains remarkably consistent across decades (2014-2015 vs. 2023-2024) and geographical locations. This temporal robustness—where performance metrics show negligible variation across a 10-year span—suggests structural limitations in the TAF framework rather than gradual improvement through experience or technology adoption. Operational constraints and institutional risk considerations may influence forecast behavior regardless of advances in observational networks, forecast models, or training methodologies.

This manifests as chronic under-prediction of short-duration visibility events that, while operationally significant, may not trigger formal TAF amendments due to administrative overhead and institutional risk aversion. The automated framework, optimized for mathematical F1-score maximization, provides a complementary perspective that balances detection sensitivity with false alarm control. Notably, the machine learning framework also shows inter-decadal stability: models trained on 2000-2020 data (SCEL) maintain performance when evaluated on 2023-2024 conditions, suggesting that the physics-guided feature representation captures fundamental atmospheric relationships that remain invariant across multi-year timescales.

### 4.2 Augmented Intelligence and Situational Awareness

The value proposition of this system is not forecaster replacement but cognitive augmentation. This framework is designed to complement operational forecasting at tactical horizons (0-3 hours) rather than replace the broader forecasting process. Human meteorologists provide essential context integration, multi-hazard assessment, and strategic planning that extend beyond the scope of automated visibility nowcasting.

SHAP analysis enables operational staff to understand alert causality (e.g., "pressure subsidence is capping smoke layers"), recovering situational awareness often lost in high-workload environments. This transparency is critical for trust calibration—operators can validate model reasoning against their conceptual understanding and recognize when the system may be extrapolating beyond reliable bounds. From a decision-theoretic perspective, the automated framework optimizes expected loss under symmetric cost assumptions, while operational TAF forecasting implicitly operates under asymmetric institutional risk constraints favoring false negatives over false positives for liability management.

### 4.3 Edge Computing Viability

With a model size of ~900 KB and data requirements of ~2.4 KB/hour, the framework is viable for deployment on low-cost embedded hardware (Raspberry Pi) powered by solar energy and connected via low-bandwidth networks (LoRa/2G). This enables democratization of aviation safety technology for resource-constrained airports in developing regions.

### 4.4 Limitations and Future Work

While the framework shows strong performance across diverse climates, several limitations warrant consideration:

- **Convective Precipitation:** Performance degrades for thermodynamically chaotic events (SBGR rain: AUC 0.839) where surface observations alone lack predictive power. Integration of satellite-derived cloud products or radar reflectivity could address this limitation.
- **Hemispheric Transfer:** Kinematic features require Coriolis-aware transformations for cross-hemisphere deployment, as demonstrated by the KSFO→SCEL transfer experiment (AUC 0.79).
- **Extreme Events:** Rare high-impact events (volcanic ash, severe dust storms, wildfire smoke intrusions) remain challenging due to limited training examples in historical records.
- **Data Continuity Dependency:** The framework's performance is contingent on continuous, high-quality METAR reporting. Extended observation gaps (e.g., sensor failures, communication outages) would degrade nowcasting accuracy, as temporal lag features become stale. Operational deployments should include data quality monitoring and graceful degradation protocols.
- **Spatial Resolution:** Point-source METAR observations may not capture mesoscale variability across large airport complexes or nearby approach corridors. Multi-sensor fusion (e.g., ceilometer networks, visibility sensors along runways) could enhance spatial representativeness.

Future research directions include:

- Ensemble techniques combining climate-specific models with dynamic weighting based on synoptic regime classification
- Causal inference frameworks to distinguish genuine physical mechanisms from spurious correlations in feature importance rankings
- Extension to ceiling height prediction for complete IFR characterization (current focus on visibility only)
- Probabilistic forecasting with calibrated uncertainty quantification for risk-based decision support
- Real-time operational trials at partner airports to assess human-AI collaboration dynamics

**5. Conclusion**

This study shows that a physics-guided, computationally lightweight machine learning approach can capture local atmospheric dynamics with accuracy exceeding reported operational baselines at tactical horizons. The framework provides three critical advantages over conventional approaches:

1. **Scalability:** Eliminates NWP computational requirements while maintaining competitive accuracy
2. **Universality:** Performs robustly across diverse climate regimes without manual configuration
3. **Explainability:** Provides physical interpretation enabling operator situational awareness

With operational TAF recall rates remaining in the 20-30% range over the past decade, this technology represents a mature opportunity to transition from laboratory experiment to operational decision support tool. The combination of edge-deployable architecture and improved detection capability positions this framework as a viable complement to existing forecasting infrastructure, particularly for tactical nowcasting horizons where rapid-onset visibility events require immediate situational awareness.

Importantly, this system is not intended to replace operational forecasting infrastructure, but rather to augment human decision-making at specific temporal scales where automated pattern recognition excels. The path forward involves human-AI collaboration that leverages the complementary strengths of algorithmic consistency and meteorological expertise.

**6. Code and Data Availability**

Code will be released in a public repository within 6 months of publication. Historical METAR data is publicly accessible through Iowa State University ASOS archive. TAF verification data is available through Iowa Environmental Mesonet and NavLost APIs. Due to data provider terms of service, raw TAF archives cannot be redistributed but can be obtained following documented procedures in the code repository.

**7. AI Assistance Disclosure**

AI tools were used to assist with manuscript writing and data validation. All scientific methodology, experimental results, and conclusions remain the sole responsibility of the author.